\documentclass[11pt,a4paper]{article}
\usepackage[hyperref]{acl2019}
\usepackage{times}
\usepackage{latexsym}
\usepackage{xspace}
\usepackage{multirow}
\usepackage{url}
\usepackage{subcaption}
\usepackage[colorinlistoftodos]{todonotes}

\aclfinalcopy 

\newcommand\BLEU{\textsc{Bleu}\xspace}
\newcommand\SacreBLEU{Sacre\textsc{Bleu}\xspace}

\title{Tagged Back-Translation}

\author{Isaac Caswell, Ciprian Chelba, David Grangier\\
  Google Research \\
  {\tt \{icaswell, ciprianchelba, grangier\}@google.com}}

\date{}

\begin{document}
\maketitle
\begin{abstract}
Recent work in Neural Machine Translation (NMT) has shown significant quality gains from noised-beam decoding during back-translation, a method to generate synthetic parallel data. We show that the main role of such synthetic noise is not to diversify the source side, as previously suggested, but simply to indicate to the model that the given source is synthetic. We propose a simpler alternative to noising techniques, consisting of tagging back-translated source sentences with an extra token. Our results on WMT outperform noised back-translation in English-Romanian and match performance on English-German, re-defining state-of-the-art in the former.
\end{abstract}

\section{Introduction}
\label{intro}

Neural Machine Translation (NMT) has made considerable progress in recent years~\cite{bahdanau2014neural, gehring2017convolutional, vaswani2017attention}. Traditional NMT has relied solely on parallel sentence pairs for training data, which can be an expensive and scarce resource. This motivates the use of monolingual data, usually more abundant \cite{lambert2011investigations}. Approaches using monolingual data for machine translation include language model fusion for both phrase-based \cite{brants2007large,koehn2009statistical} and neural MT \cite{gulcehre2015using,gulcehre2017integrating}, back-translation \cite{sennrich2016improving,poncelas2018investigating}, unsupervised machine translation \cite{lample2017unsupervised, artetxe2018unsupervised}, dual learning  \cite{cheng2016semisupervised, he2016dual, xia2017dual}, and multi-task learning \cite{domhan2017using}.

We focus on back-translation (BT), which, despite its simplicity, has thus far been the most effective technique \cite{sennrich2017edinburgh,ha2017effective,garcia2017lium}. Back-translation entails training an intermediate target-to-source model on genuine bitext, and using this model to translate a large monolingual corpus from the target into the source language. This allows training a source-to-target model on a mixture of genuine parallel data and synthetic pairs from back-translation.

We build upon \newcite{edunov2018understanding} and \newcite{imamura2018enhancement}, who investigate BT at the scale of hundreds of millions of sentences. Their work studies different decoding/generation
methods for back-translation: in addition to regular beam search, they consider sampling and adding noise to the one-best hypothesis produced by beam search. They show that sampled BT and noised-beam BT significantly outperform standard BT, and attribute this success to increased source-side diversity (sections 5.2 and 4.4).

Our work investigates noised-beam BT (NoisedBT) and questions the role noise is playing. Rather than increasing source diversity, our work instead suggests that the performance gains come simply from signaling to the model that the source side is back-translated, allowing it to treat the synthetic parallel data differently than the natural bitext. We hypothesize that BT introduces both helpful signal (strong target-language signal and weak cross-lingual signal) and harmful signal (amplifying the biases of machine translation). Indicating to the model whether a given training sentence is back-translated should allow the model to separate the helpful and harmful signal. 

To support this hypothesis, we first demonstrate that the permutation and word-dropping noise used by \citet{edunov2018understanding} do not improve or significantly degrade NMT accuracy, corroborating that noise might act as an indicator that the source is back-translated, without much loss in mutual information between the source and target. We then train models on WMT English-German (EnDe) without BT noise, and instead explicitly tag the synthetic data with a reserved token. We call this technique ``Tagged Back-Translation" (TaggedBT). These models achieve equal to slightly higher performance than the noised variants. We repeat these experiments with WMT English-Romanian (EnRo), where NoisedBT underperforms standard BT and TaggedBT improves over both techniques. We demonstrate that TaggedBT also allows for effective iterative back-translation with EnRo, a technique which saw quality losses when applied with standard back-translation.

To further our understanding of TaggedBT, we investigate the biases encoded in models by comparing the entropy of their attention matrices, and look at the attention weight on the tag. We conclude by investigating the effects of the back-translation tag at decoding time.


\section{Related Work}

This section describes prior work exploiting target-side monolingual data and discusses related work tagging NMT training data.

\subsection{Leveraging Monolingual Data for NMT}

Monolingual data can provide valuable information to improve 
translation quality. Various methods for using target-side LMs have proven effective for  NMT~\cite{he2016improved,gulcehre2017integrating}, but have tended to be less successful than back-translation -- for example, \citet{gulcehre2017integrating} report under +0.5 \BLEU over their baseline on EnDe newstest14, whereas \citet{edunov2018understanding} report over +4.0 \BLEU on the same test set. Furthermore, there is no straighforward way to incorporate source-side monolingual data into a neural system with a LM.

Back-translation was originally introduced for phrase-based systems~\cite{bertoldi2009domain,bojar2011improving}, but flourished in NMT after work by~\newcite{sennrich2016improving}. Several approaches have looked into iterative forward- and BT experiments (using source-side monolingual data), including \newcite{cotterell2018explaining}, \newcite{hoang2018iterative}, and \newcite{niu2018bidirectional}. Recently, iterative back-translation in both directions
has been devised has a way to address unsupervised machine translation~\cite{lample2018phrase,artetxe2018unsupervised}.

Recent work has focused on the importance of diversity and complexity in synthetic training data. \citet{fadaee2018backtranslation} find that BT benefits difficult-to-translate words the most, and select from the back-translated corpus by oversampling words with high prediction loss. \citet{imamura2018enhancement} argue that in order for BT to enhance the encoder, it must have a more diverse source side, and sample several back-translated source sentences for each monolingual target sentence. Our work follows most closely \citet{edunov2018understanding}, who investigate alternative decoding schemes for BT. Like \citet{imamura2018enhancement}, they argue that BT through beam or greedy decoding leads to an overly regular domain on the source side, which poorly represents the diverse distribution of natural text.

Beyond the scope of this work, we briefly mention alternative techniques leveraging
monolingual data, like forward translation~\cite{ueffing2007semi,kim-rush-2016-sequence}, 
or source copying~\cite{currey2017}.

\subsection{Training Data Tagging for NMT}

Tags have been used for various purposes in NMT. Tags on the source sentence can indicate the target language in multi-lingual models~\cite{johnson2016google}. \citet{yamagishi2016controlling} use tags in a similar fashion to control the formality of a translation from English to Japanese. \citet{kuczmarski2018gender} use tags to control gender in translation. Most relevant to our work, \citet{kobus2016controlling} use tags to mark source sentence domain in a multi-domain setting.

\section{Experimental Setup}

This section presents our datasets, evaluation protocols and model architectures. It also describes our back-translation procedure, as well as noising and tagging strategies. 

\subsection{Data}
We perform our experiments on WMT18 EnDe bitext, WMT16 EnRo bitext, and WMT15 EnFr bitext respectively. We use WMT Newscrawl for monolingual data (2007-2017 for De, 2016 for Ro, 2007-2013 for En, and 2007-2014 for Fr). For bitext, we filter out empty sentences and sentences longer than 250 subwords. We remove pairs whose whitespace-tokenized length ratio is greater than 2. This results in about 5.0M pairs for EnDe, and 0.6M pairs for EnRo. We do not filter the EnFr bitext, resulting in 41M sentence pairs.

For monolingual data, we deduplicate and filter sentences with more than 70 tokens or 500 characters. Furthermore, after back-translation, we remove any sentence pairs where the back-translated source is longer than 75 tokens or 550 characters. This results in 216.5M sentences for EnDe, 2.2M for EnRo, 149.9M for RoEn, and 39M for EnFr. For monolingual data, all tokens are defined by whitespace tokenization, not wordpieces.

The DeEn model used to generate BT data has 28.6 \SacreBLEU on newstest12, the RoEn model used for BT has a test \SacreBLEU of 31.9 (see Table~\ref{enro}.b), and the FrEn model used to generate the BT data has 39.2 \SacreBLEU on newstest14.

\subsection{Evaluation}

We rely on \BLEU score~\cite{papineni2002bleu} as our evaluation metric.

While well established, any slight difference in post-processing and \BLEU computation can have a dramatic impact on output values \cite{post2018call}. For example, \citet{lample2019cross} report 33.3 \BLEU on EnRo using unsupervised NMT, which at first seems comparable to our reported 33.4 \SacreBLEU from iterative TaggedBT. However, when we use their preprocessing scripts and evaluation protocol, our system achieves 39.2 \BLEU on the same data, which is close to 6 points higher than the same model evaluated by \SacreBLEU.

We therefore report strictly \SacreBLEU\footnote{BLEU + case.mixed + lang.LANGUAGE\_PAIR + numrefs.1 + smooth.exp + test.SET + tok.13a + version.1.2.15}, using the reference implementation from ~\newcite{post2018call}, which aims to standardize \BLEU evaluation.

\subsection{Architecture}
We use the transformer-base and transformer-big architectures~\cite{vaswani2017attention} implemented in {\it lingvo}~\cite{shen2019lingvo}. Transformer-base is used for the bitext noising experiments and the EnRo experiments, whereas the transformer-big is used for the EnDe tasks with BT. Both use a vocabulary of 32k subword units. As an alternative to the checkpoint averaging used in \newcite{edunov2018understanding}, we train with exponentially weighted moving average (EMA) decay with weight decay parameter $\alpha=0.999$~\cite{buduma2017fundamentals}.

Transformer-base models are trained on 16 GPUs with synchronous gradient updates and per-gpu-batch-size of 4,096 tokens, for an effective batch size of 64k tokens/step. Training lasts 400k steps, passing over 24B tokens. For the final EnDe TaggedBT model, we train transformer-big similarly but on 128 GPUs, for an effective batch size of 512k tokens/step. A training run of 300M steps therefore sees about 150B tokens. We pick checkpoints with newstest2012 for EnDe and newsdev2016 for EnRo. 

\subsection{Noising} \label{section:noising}
We focused on noised beam BT, the most effective noising approach according to \citet{edunov2018understanding}. Before training, we noised the decoded data \cite{lample2017unsupervised} by applying 10\% word-dropout, 10\% word blanking, and a 3-constrained permutation (a permutation such that no token moves further than 3 tokens from its original position). We refer to data generated this way as NoisedBT.  Additionally, we experiment using only the 3-constrained permutation and no word dropout/blanking, which we abbreviate as P3BT. 

\subsection{Tagging} \label{section:tagging}
We tag our BT training data by prepending a reserved token to the input sequence, which is then treated in the same way as any other token. We also experiment with both noising and tagging together, which we call Tagged Noised Back-Translation, or TaggedNoisedBT. This consists simply of prepending the $<$BT$>$ tag to each noised training example.

An example training sentence for each of these set-ups can be seen in Table~\ref{noise_example}. We do not tag the bitext, and always train on a mix of back-translated data and (untagged) bitext unless explicitly stated otherwise.

\begin{table}[ht]
\small
\centering
\begin{tabular}{l|l}
 Noise type & Example sentence  \\
 \hline
[no noise] & Raise the child, love the child. \\
P3BT & child Raise the, love child the. \\
NoisedBT & Raise child \underline{\hspace{5mm}}  love child, the. \\
TaggedBT & $<$BT$>$ Raise the child, love the child. \\
TaggedNoisedBT & $<$BT$>$ Raise, the child the \underline{\hspace{5mm}}   love. \\
\end{tabular}
\caption{Examples of the five noising settings examined in this paper \label{noise_example}}
\end{table}


\section{Results}

This section studies the impact of training data noise on translation quality, and then presents our results with TaggedBT on EnDe and EnRo.

\subsection{Noising Parallel Bitext}

We first show that noising EnDe bitext sources does not seriously impact the translation quality of the transformer-base baseline.
For each sentence pair in the corpus, we flip a coin and noise the source sentence with probability $p$. We then train a model from scratch on this partially noised dataset. Table \ref{noise_bitext} shows results for various values of $p$.
Specifically, it presents the somewhat unexpected finding that even when noising 100\% of the source bitext (so the model has never seen well-formed English), \BLEU on well-formed test data only drops by 2.5.

This result prompts the following line of reasoning about the role of noise in BT: (i) By itself, noising does not add meaningful signal (or else it would improve performance); (ii) It also does not damage the signal much; (iii) In the context of back-translation, the noise could therefore signal whether a sentence were back-translated, without significantly degrading performance.

\begin{table}[ht]
\small
\centering
\begin{tabular}{l|cc}
          & \multicolumn{2}{c}{\SacreBLEU} \\
\% noised & Newstest '12 & Newstest '17 \\
 \hline\hline
0\% & 22.4 & 28.1 \\
20\% & 22.4 & 27.9 \\
80\% & 21.5  & 27.0 \\
100\% & 21.2 & 25.6 \\
\end{tabular}
\caption{\SacreBLEU degradation as a function of the proportion of bitext data that is noised. \label{noise_bitext}}
\end{table}

\subsection{Tagged Back-Translation for EnDe}
We compare the results of training on a mixture of bitext and a random sample of 24M back-translated sentences in Table~\ref{en_de_bt}.a, for the various set-ups of BT described in sections \ref{section:noising} and \ref{section:tagging}. Like \newcite{edunov2018understanding}, we confirm that BT improves over bitext alone, and noised BT improves over standard BT by about the same margin. All methods of marking the source text as back-translated (NoisedBT, P3BT, TaggedBT, and TaggedNoisedBT) perform about equally, with TaggedBT having the highest average \BLEU by a small margin. Tagging and noising together (TaggedNoisedBT) does not improve over either tagging or noising alone, supporting the conclusion that tagging and noising are not orthogonal signals but rather different means to the same end.  

\begin{table*}[ht]
\small
\centering
\begin{tabular}{l|cccccccccc}
\multicolumn{4}{c}{a. Results on 24M BT Set}\\
Model & AVG 13-18 & 2010 & 2011 & 2012 & 2013 & 2014 & 2015 & 2016 & 2017 & 2018 \\
 \hline\hline
Bitext & 32.05  & 24.8 & 22.6 & 23.2 & 26.8 & 28.5 & 31.1 & 34.7 & 29.1 & 42.1 \\                                                         
\hline
BT & 33.12 & 24.7 & 22.6 & 23.5 & 26.8 & 30.8 & 30.9 & 36.1 & 30.6 & 43.5 \\
\hline
NoisedBT & 34.70  & 26.2 & \textbf{23.7 }& \textbf{24.7} & \textbf{28.5 }& 31.3 & 33.1 & 37.7 & \textbf{31.7 }& \textbf{45.9} \\
P3BT & 34.57  & 26.1 & 23.6 & 24.5 & 28.1 & 31.8 & 33.0 & 37.4 & 31.5 & 45.6 \\
TaggedBT & \textbf{34.83 } & \textbf{26.4} & 23.6 & 24.5 & 28.1 & \textbf{32.1} & \textbf{33.4} & \textbf{37.8} &\textbf{ 31.7} & \textbf{45.9} \\
TaggedNoisedBT & 34.52 &  26.3 & 23.4 & 24.6 & 27.9 & 31.4 & 33.1 & 37.4 &\textbf{ 31.7} & 45.6 \\
\hline
BT alone & 31.20 & 23.5 & 21.2 & 22.7 & 25.2 & 29.3 & 29.4 & 33.7 & 29.1 & 40.5 \\
NoisedBT alone & 30.28 & 23.2 & 21.0 & 22.1 & 24.6 & 28.4 & 28.2 & 33.0 & 28.1 & 39.4 \\
\hline
Noised(BT + Bitext) & 32.07 &  24.2 & 22.1 & 23.5 & 26.2 & 29.7 & 30.1 & 35.1 & 29.4 & 41.9 \\
+ Tag on BT  & 33.53 & 25.5 & 22.8 & 24.5 & 27.6 & 30.3 & 31.9 & 36.9 & 30.4 & 44.1 \\
\multicolumn{4}{c}{}\\
\multicolumn{4}{c}{b. Results on 216M  BT Set}\\
Model & AVG 13-18 & 2010 & 2011 & 2012 & 2013 & 2014 & 2015 & 2016 & 2017 & 2018 \\
\hline\hline
\citet{edunov2018understanding} & 35.28 &  &  & 25.0 & \textbf{29.0} & \textbf{33.8} & 34.4 & 37.5 & \textbf{32.4} & 44.6 \\
NoisedBT & 35.17 & \textbf{26.7} & 24.0 & \textbf{25.2 }& 28.6 & 32.6 & 33.9 & 38.0 & 32.2 & 45.7 \\
TaggedBT & \textbf{35.42} & 26.5 & \textbf{24.2 }& \textbf{25.2} & 28.7 & 32.8 & \textbf{34.5} & \textbf{38.1 }& \textbf{32.4} & \textbf{46.0} \\
\end{tabular}
\caption{\SacreBLEU on Newstest EnDe for different types of noise, with back-translated data either sampled down to 24M or using the full set of 216M sentence pairs. \label{en_de_bt}}
\end{table*}

Table~\ref{en_de_bt}.b verifies our result at scale applying TaggedBT on the full BT dataset (216.5M sentences), upsampling the bitext so that each batch contains an expected 20\% of bitext. As in the smaller scenario, TaggedBT matches or slightly out-performs NoisedBT, with an advantage on seven test-sets and a disadvantage on one. We also compare our results to the best-performing model from \newcite{edunov2018understanding}. Our model is on par with or slightly superior to their result\footnote{\SacreBLEU for the WMT-18 model at \url{github.com/pytorch/fairseq}}, out-performing it on four test sets and under-performing it on two, with the largest advantage on Newstest2018 (+1.4 \BLEU).



As a supplementary experiment, we consider training only on BT data, with no bitext. We compare this to training only on NoisedBT data. If noising in fact increases the quality or diversity of the data, one would expect the NoisedBT data to yield higher performance than training on unaltered BT data, when in fact it has about 1 \BLEU lower performance (Table \ref{en_de_bt}.a, ``BT alone" and ``NoisedBT alone").

We also compare NoisedBT versus TaggedNoisedBT in a set-up where the bitext itself is noised. In this scenario, the noise can no longer be used by the model as an implicit tag to differentiate between bitext and synthetic BT data, so we expect the TaggedNoisedBT variant to perform better than NoisedBT by a similar margin to NoisedBT's improvement over BT in the unnoised-bitext setting. The last sub-section of Table \ref{en_de_bt}.a confirms this.

\subsection{Tagged Back-Translation for EnRo}

We repeat these experiments for WMT EnRo (Table \ref{enro}). This is a much lower-resource task than EnDe, and thus can benefit more from monolingual data. In this case, NoisedBT is actually harmful, lagging standard BT by -0.6 \BLEU. TaggedBT closes this gap and passes standard BT by +0.4 \BLEU, for a total gain of +1.0 \BLEU over NoisedBT.
\begin{table}[ht]
\small
\centering
\begin{tabular}{l|cc}
\multicolumn{3}{c}{a. Forward models (EnRo)}  \\
Model & dev & test \\
 \hline\hline
\citet{gehring2017convolutional} &  & \textit{29.9}  \\
Sennrich 2016 (BT) & \textit{29.3} &\textit{ 28.1} \\
\hline
bitext & 26.5 & 28.3 \\
\hline
BT & 31.6 & 32.6 \\
NoisedBT & 29.9 & 32.0 \\
TaggedBT & 30.5 & 33.0 \\
It.-3 BT & 31.3 & 32.8  \\
It.-3 NoisedBT & 31.2	& 32.6 \\
It.-3 TaggedBT & 31.4	& \textbf{33.4} \\
\multicolumn{3}{c}{}  \\
\multicolumn{3}{c}{b. Reverse models (RoEn)}  \\
Model & dev & test \\
\hline\hline
bitext & 32.9 & 31.9 \\
It.-2 BT & 39.5 & \textbf{37.3} \\
\end{tabular}
\caption{Comparing \SacreBLEU scores for different flavors of BT for WMT16 EnRo. Previous works' scores are reported in italics as they use \texttt{detok.multi-bleu} instead of \SacreBLEU, so are not guaranteed to be comparable. In this case, however, we do see identical \BLEU on our systems when we score them with \texttt{detok.multi-bleu}, so we believe it to be a fair comparison. \label{enro}}
\end{table}

\subsection{Tagged Back-Translation for EnFr}
We performed a minimal set of experiments on WMT EnFr, which are summarized in Table \ref{enfr}. This is a much higher-resource language pair than either EnRo or EnDe, but \citet{edunov2018understanding} demonstrate that noised BT (using sampling) can still help in this set-up. In this case, we see that BT alone hurts performance compared to the strong bitext baseline, but NoisedBT indeed surpasses the bitext model. TaggedBT  out-performs all other methods, beating NoisedBT by an average of +0.3 \BLEU over all test sets.

It is worth noting that our numbers are lower than those reported by \citet{edunov2018understanding} on the years they report (36.1, 43.8, and 40.9 on 2013, 2014, and 2015 respectively). We did not investigate this result. We suspect that this is an error/inoptimlaity in our set-up, as we did not optimize these models, and ran only one experiment for each of the four set-ups. Alternately, sampling could outperform noising in the large-data regime.

\begin{table*}[ht]
\small
\centering
\begin{tabular}{l|ccccccccc}
Model & Avg &   2008 &  2009 &  2010 &  2011 &  2012 &  2013 &  2014 &  2015 \\
 \hline\hline
Bitext & 32.8 &  26.3 &  28.8 &  32.0 &  32.9 &  30.1 &  33.5 &  40.6 &  38.4 \\
BT & 29.2 &  22.2 &  27.3 &  28.8 &  29.3 &  27.9 &  30.7 &  32.6 &  34.8 \\
NoisedBT & 33.8 &  26.8 &  29.9 &  33.4 &  \textbf{33.9} &  \textbf{31.3} &  34.3 &  42.3 &  38.8 \\
TaggedBT & \textbf{34.1} & \textbf{ 27.0 }&  \textbf{30.0} &  \textbf{33.6} &  \textbf{33.9} &  31.2 &  \textbf{34.4} &  \textbf{42.7} &  \textbf{39.8} \\
\hline
\end{tabular}
\caption{Results on WMT15 EnFr, with bitext, BT, NoisedBT, and TaggedBT. \label{enfr}}
\end{table*}

\subsection{Iterative Tagged Back-Translation}
We further investigate the effects of TaggedBT by performing one round of iterative back-translation \cite{cotterell2018explaining,hoang2018iterative,niu2018bidirectional}, and find another difference between the different varieties of BT: NoisedBT and TaggedBT allow the model to bootstrap improvements from an improved reverse model, whereas standard BT does not. This is consistent with our argument that data tagging allows the model to extract information out of each data set more effectively.

For the purposes of this paper we call a model trained with standard back-translation an \textbf{Iteration-1 BT model}, where the back-translations were generated by a model trained only on bitext. We inductively define the \textbf{Iteration-k BT model} as that model which is trained on BT data generated by an \mbox{\textbf{Iteration-(k-1) BT}} model, for $k>1$. Unless otherwise specified, any BT model mentioned in this paper is an Iteration-1 BT model.

We perform these experiments on the English-Romanian dataset, which is smaller and thus better suited for this computationally expensive process. We used the (Iteration-1) TaggedBT model to generate the RoEn back-translated training data. Using this we trained a superior RoEn model, mixing 80\% BT data with 20\% bitext. Using this Iteration-2 RoEn model, we generated new EnRo BT data, which we used to train the Iteration-3 EnRo models. \SacreBLEU scores for all these models are displayed in Table~\ref{enro}.

We find that the iteration-3 BT models improve over their Iteration-1 counterparts only for NoisedBT (+1.0 \BLEU, dev+test avg) and TaggedBT (+0.7 \BLEU, dev+test avg), whereas the Iteration-3 BT model shows no improvement over its Iteration-1 counterpart (-0.1 \BLEU, dev+test avg). In other words, both techniques that (explicitly or implicitly) tag synthetic data benefit from iterative BT. We speculate that this separation of the synthetic and natural domains allows the model to bootstrap more effectively from the increasing quality of the back-translated data while not being damaged by its quality issues, whereas the simple BT model cannot make this distinction, and is equally ``confused" by the biases in higher or lower-quality BT data.

An identical experiment with EnDe did not see either gains or losses in \BLEU from iteration-3 TaggedBT. This is likely because there is less room to bootstrap with the larger-capacity model. This said, we do not wish to read too deeply into these results, as the effect size is not large, and neither is the number of experiments. A more thorough suite of experiments is warranted before any strong conclusions can be made on the implications of tagging on iterative BT.


\section{Analysis}

In an attempt to gain further insight into TaggedBT as it compares with standard BT or NoisedBT, we examine attention matrices in the presence of the back translation tag and measure the impact of the tag at decoding time.

\begin{figure*}[!ht]
\begin{subfigure}{.5\textwidth}
\begin{subfigure}{1.0\textwidth}
  \centering
  \includegraphics[width=.8\linewidth]{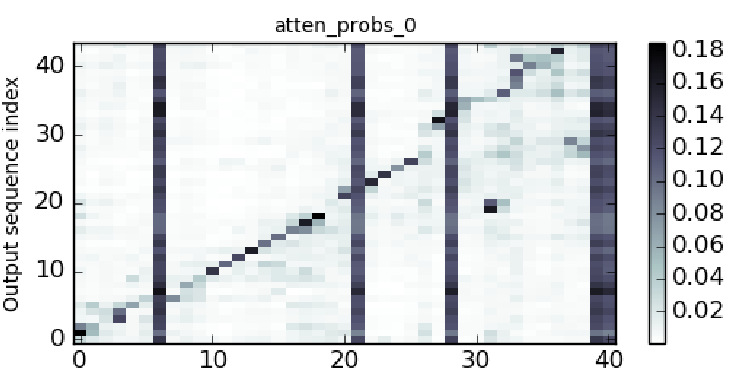}
  \caption{EnDe BT}
  \label{pretty-attention:ende-bt}
  \end{subfigure}
\begin{subfigure}{1.0\textwidth}
  \centering
  \includegraphics[width=.8\linewidth]{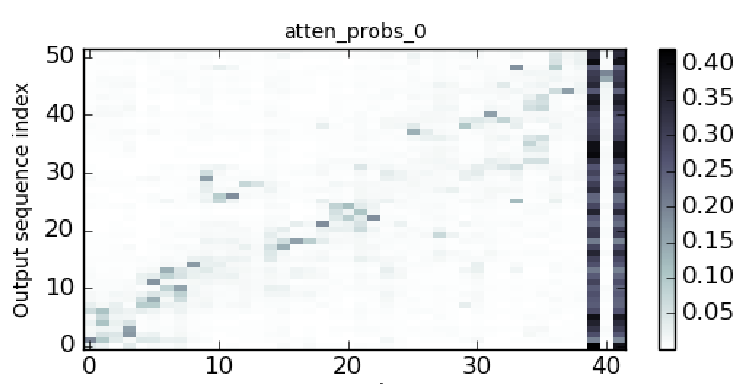}
  \caption{EnDe NoisedBT}
  \label{pretty-attention:ende-nbt}
  \end{subfigure}
\begin{subfigure}{1.0\textwidth}
  \centering
  \includegraphics[width=.8\linewidth]{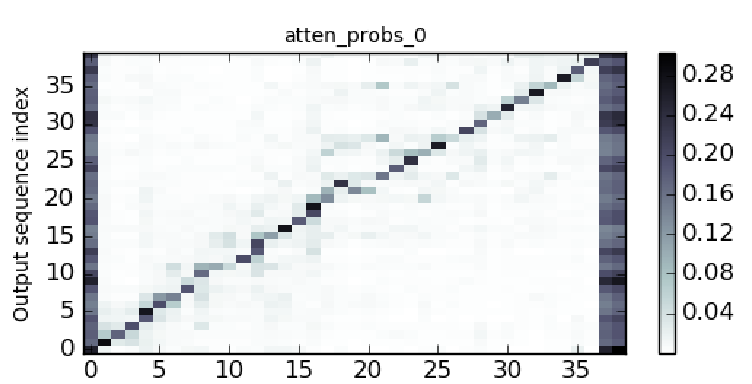}
  \caption{EnDe TaggedBT}
  \label{pretty-attention:ende-tbt}
  \end{subfigure}
\end{subfigure}%
\begin{subfigure}{.5\textwidth}
\begin{subfigure}{1.0\textwidth}
  \centering
  \includegraphics[width=.8\linewidth]{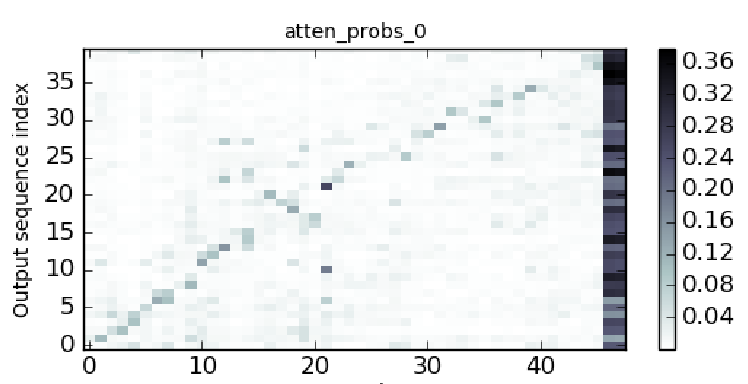}
  \caption{EnRo BT}
  \label{pretty-attention:enro-bt}
  \end{subfigure}
\begin{subfigure}{1.0\textwidth}
  \centering
  \includegraphics[width=.8\linewidth]{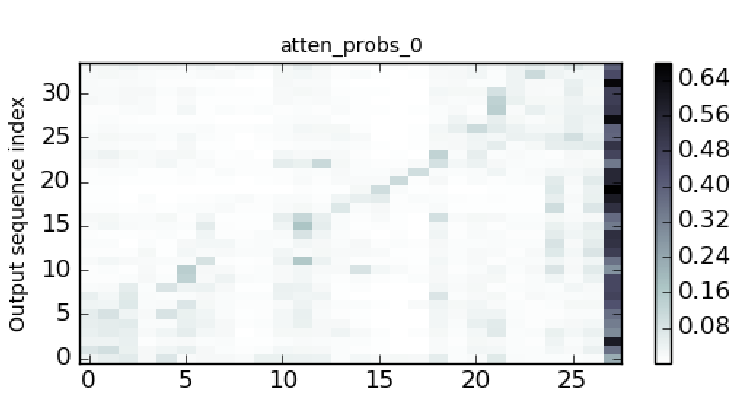}
  \caption{EnRo NoisedBT}
  \label{pretty-attention:enro-nbt}
  \end{subfigure}
\begin{subfigure}{1.0\textwidth}
  \centering
  \includegraphics[width=.8\linewidth]{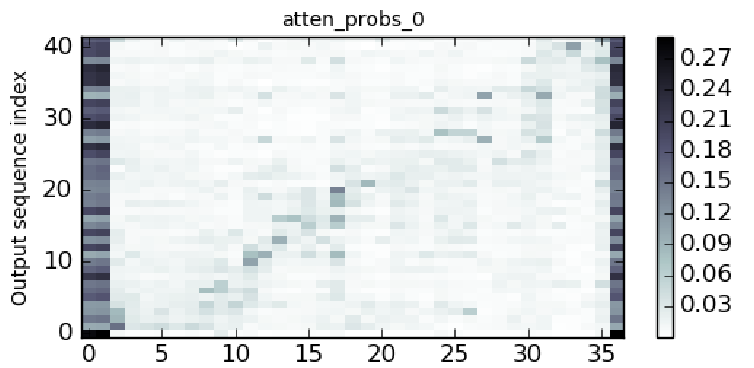}
  \caption{EnRo TaggedBT}
  \label{pretty-attention:enro-tbt}
  \end{subfigure}
\end{subfigure}
\caption{Comparison of attention maps at the first encoder layer for a random training example for BT (row 1), NoisedBT (row 2), and TaggedBT (row 3), for both EnDe (col 1) and EnRo (col 2). Note the heavy attention on the tag (position 0 in row 3), and the diffuse attention map learned by the NoiseBT models. These are the models from Table~\ref{en_de_bt}.a}
\label{pretty-attention}
\end{figure*}

\subsection{Attention Entropy and Sink-Ratio}

To understand how the model treats the tag and what biases it learns from the data, we investigate the entropy of the attention probability distribution, as well as the attention captured by the tag.

We examine decoder attention (at the top layer) on the first source token. We define Attention Sink Ratio for index $j$ ($\textrm{ASR}_j$) as the averaged attention 
over the $j$th token, normalized by uniform attention, i.e.
$$
\textrm{ASR}_j(x, \hat y) = \frac{1}{|\hat y|} \sum_{i = 1}^{ | \hat y | } \frac{\alpha_{ij}}{\tilde{\alpha}} 
$$
where $\alpha_{ij}$ is the attention value for target token $i$ in hypothesis $\hat y$ over source token $j$ and $\tilde{\alpha} = \frac{1}{|x|}$ corresponds to uniform attention.
We examine ASR on text that has been noised and/or tagged (depending on the model), to understand how BT sentences are treated during training. For the tagged variants, there is heavy attention on the tag when it is present (Table \ref{attn_stats}), indicating that the model relies on the information signalled by the tag. 

Our second analysis probes word-for-word translation bias through the average source-token entropy of the attention probability model when decoding natural text. Table~\ref{attn_stats} reports the average length-normalized Shannon entropy:
$$
\tilde \textrm{H}(x, \hat y) = - \frac{1}{|\hat y|} \sum_{i=1}^{| \hat y|} \frac{1}{\log |x|} \sum_{j=1}^{|x|}\alpha_{ij} \log(\alpha_{ij}) 
$$
The entropy of the attention probabilities from the model trained on BT data is the clear outlier. This low entropy corresponds to a concentrated attention matrix, which we observed to be concentrated on the diagonal (See Figure~\ref{pretty-attention:ende-bt} and \ref{pretty-attention:enro-bt}). This could
indicate the presence of word-by-word translation, a consequence of the harmful part of the signal from back-translated data. The entropy on parallel data from the NoisedBT model is much higher, corresponding to more diffuse attention, which we see in Figure~\ref{pretty-attention:ende-nbt} and ~\ref{pretty-attention:enro-nbt}. In other words, the word-for-word translation biases in BT data, that were incorporated into the BT model, have been manually undone by the noise, so the model's understanding of how to decode parallel text is not corrupted. We see that TaggedBT leads to a similarly high entropy, indicating the model has learnt this without needing to manually ``break" the literal-translation bias. As a sanity check, we see that the entropy of the P3BT model's attention is also high, but is lower than that of the NoisedBT model, because P3 noise is less destructive. The one surprising entry on this table is probably the low entropy of the TaggedNoisedBT. Our best explanation is that TaggedNoisedBT puts disproportionately high attention on the sentence-end token, with 1.4x the $\textrm{ASR}_{|x|}$ that TaggedBT has, naturally leading to lower entropy.

\begin{table}[ht]
\small
\centering
\begin{tabular}{l|c|c|c}
Model & $\textrm{ASR}_0$ & $\textrm{ASR}_{|x|}$ & $\tilde \textrm{H}$ \\
       \hline\hline
       Bitext baseline   & 0.31 & 10.21 & 0.504 \\
       BT & 0.28 & 10.98 &  \textbf{0.455} \\
       P3BT & 0.37 & 7.66 &  0.558 \\
       NoisedBT & 1.01 & 3.96 & 0.619 \\
       TaggedBT  & \textbf{5.31} & 5.31  & 0.597 \\
       TaggedNoisedBT  & \textbf{7.33} & 7.33 & 0.491 \\
\end{tabular}
\small
\caption{Attention sink ratio on the first and last token and entropy (at decoder layer 5) for the models in Table~\ref{en_de_bt}.a, averaged over all sentences in newstest14. For ASR, data is treated as if it were BT (noised and/or tagged, resp.), whereas for entropy the natural text is used. Outliers discussed in the text are bolded.  \label{attn_stats}}
\end{table}


\subsection{Decoding with and without a tag}

\begin{table*}[ht]
\small
\centering
\begin{tabular}{l|cccccccccc}
Model & Decode type & AVG 13-17 & 2010 & 2011 & 2012 & 2013 & 2014 & 2015 & 2016 & 2017  \\
 \hline\hline
TaggedBT & \multicolumn{1}{l}{standard} & \textbf{33.24} & 26.5 & \textbf{24.2} & \textbf{25.2} & \textbf{28.7} & \textbf{32.8} &\textbf{ 34.5 }& \textbf{38.1} & \textbf{32.4} \\
 & \multicolumn{1}{l}{as BT (tagged)}  & 30.30 & 24.3 & 22.2 & 23.4 & 26.6 & 30.0 & 30.5 & 34.2 & 30.2  \\
NoisedBT & \multicolumn{1}{l}{standard} & 33.06 & \textbf{26.7} & 24.0 & \textbf{25.2} & 28.6 & 32.6 & 33.9 & 38.0 & 32.2 \\
 & \multicolumn{1}{l}{as BT (noised)} & 10.66 & 8.1 & 6.5 & 7.5 & 8.2 & 11.1 & 10.0 & 12.7 & 11.3 \\
\hline
\end{tabular}
\caption{Comparing standard decoding with decoding as if the input were back-translated data, meaning that it is tagged (for the TaggedBT model) or noised (for the NoisedBT model) . \label{tagged_decode}}
\end{table*}

\begin{table*}[ht]
\small
\centering
\begin{tabular}{l|ll}
Model & Decode type & Example \\
 \hline\hline
TaggedBT & \multicolumn{1}{l|}{standard} & Wie der Reverend Martin Luther King Jr. vor f{\" u}nfzig Jahren sagte: \\
 & \multicolumn{1}{l|}{as-if-BT (tagged)}  & Wie sagte der Reverend Martin Luther King jr. Vor f{\" u}nfzig Jahren: \\
NoisedBT & \multicolumn{1}{l|}{standard} & Wie der Reverend Martin Luther King Jr. vor f{\" u}nfzig Jahren sagte: \\
 & \multicolumn{1}{l|}{as-if-BT (noised)} & Als Luther King Reverend  \underline{\hspace{5mm}} Jr. vor f{\" u}nfzig Jahren:\\
\hline
 Source & & As the Reverend Martin Luther King Jr. said fifty years ago:\\
 Reference & & Wie Pastor Martin Luther King Jr. vor f{\" u}nfzig Jahren sagte: \\
\end{tabular}
\small
\caption{Example decodes from newstest2015 for decoding in standard and ``as-if-BT" varieties. Here, NoisedBT and TaggedBT produce equivalent outputs with standard decoding; TaggedBT produces less natural output with tagged input; and NoisedBT produces a low-quality output with noised input. \label{decode_examples}}                                                                     
\end{table*}

In this section we look at what happens when we decode with a model on newstest data as if it were back-translated. This means that for the TaggedBT model we tag the true source, and for the NoisedBT model, we noise the true source. These ``as-if-BT" decodings contrast with ``standard decode", or decoding with the true source. An example sentence from newstest2015 is shown in Table~\ref{decode_examples}, decoded by both models both in the standard fashion and in the ``as-if-BT" fashion. The \BLEU scores of each decoding method are presented in Table~ \ref{tagged_decode}.

The noised decode -- decoding newstest sentences with the NoisedBT model after noising the source -- yields poor performance. This is unsurprising given the severity of the noise model used (recall Table~\ref{noise_example}). The tagged decode, however, yields only somewhat lower performance than the standard decode on the same model (-2.9\BLEU on average). There are no clear reasons for this quality drop -- the model correctly omits the tag in the outputs, but simply produces slightly lower quality hypotheses. The only noticeable difference in decoding outputs between the two systems is that the tagged decoding produces about double the quantity of English outputs (2.7\% vs. 1.2\%, over newstest2010-newstest2017, using a language ID classifier).

That the tagged-decode \BLEU is still quite reasonable tells us that the model has not simply learned to ignore the source sentence when it encounters the input tag, suggesting that the $p(y|\textrm{BT}(x))$ signal is still useful to the model, as \newcite{sennrich2016improving} also demonstrated. The tag might then be functioning as a domain tag, causing the model to emulate the domain of the BT data -- including both the desirable target-side news domain and the MT biases inherent in BT data.

To poke at the intuition that the quality drop comes in part from emulating the NMT biases in the synthetic training data, we probe a particular shortcoming of NMT: copy rate. We quantify the copy rate with the unigram overlap between source and target as a percentage of tokens in the target side, and compare those statistics to the bitext and the back-translated data (Table \ref{st_overlap}). We notice that the increase in unigram overlap with the tagged decode corresponds to the increased copy rate for the back-translated data (reaching the same value of 11\%), supporting the hypothesis that the tag helps the model separate the domain of the parallel versus the back-translated data. Under this lens, quality gains from TaggedBT/NoisedBT could be re-framed as transfer learning from a multi-task set-up, where one task is to translate simpler ``translationese" \cite{gellerstam1986translationese, freitag2019repair} source text, and the other is to translate true bitext.

\begin{table}[ht]
\small
\centering
\begin{tabular}{l|c}
Data & src-tgt unigram overlap \\
 \hline\hline
TaggedBT (standard decode) & 8.9\% \\
TaggedBT (tagged decode) & 10.7\%\\
Bitext & 5.9\%\\
BT Data & 11.4 \%\\
\hline
\end{tabular}
\caption{Source-target overlap for both back-translated data with decoding newstest as if it were bitext or BT data. Model decodes are averaged over newstest2010-newstest2017. \label{st_overlap}}
\end{table}

\section{Negative Results}
In addition to tagged back-translation, we tried several tagging-related experiments that did not work as well. We experimented with tagged forward-translation (TaggedFT), and found that the tag made no substantial difference, often lagging behind untagged forward-translation (FT) by a small margin ($\sim$ 0.2 \BLEU). For EnDe, (Tagged)FT underperformed the bitext baseline; for EnRo, (Tagged)FT performed about the same as BT. Combining BT and FT had additive effects, yielding results slightly higher than iteration-3 TaggedBT (Table \ref{enro}), at 33.9 \SacreBLEU on test; but tagging did not help in this set-up. We furthermore experimented with year-specific tags on the BT data, using a different tag for each of the ten years of newscrawl. The model trained on these data performed identically to the normal TaggedBT model. Using this model we replicated the ``as-if-bt" experiments from Table \ref{decode_examples} using year-specific tags, and although there was a slight correlation between year tag and that year's dataset, the standard-decode still resulted in the highest \BLEU.

\section{Conclusion}
In this work we develop TaggedBT, a novel technique for using back-translation in the context of NMT, which improves over the current state-of-the-art method of Noised Back-Translation, while also being simpler and more robust. We demonstrate that while Noised Back-Translation and standard Back-Translation are more or less effective depending on the task (low-resource, mid-resource, iterative BT), TaggedBT performs well on all tasks.


On WMT16 EnRo, TaggedBT improves on vanilla BT by 0.4 \BLEU. Our best \BLEU score of 33.4 \BLEU, obtained using Iterative TaggedBT, shows a gain of +3.5 \BLEU over the highest previously published result on this test-set that we are aware of. We furthermore match or out-perform the highest published results we are aware of on WMT EnDe that use only back-translation, with higher or equal \BLEU on five of seven test sets.

In addition, we conclude that noising in the context of back-translation acts merely as an indicator to the model that the source is back-translated, allowing the model to treat it as a different domain and separate the helpful signal from the harmful signal. We support this hypothesis with experimental results showing that heuristic noising techniques like those discussed here, although they produce text that may seem like a nigh unintelligible mangling to humans, have a relatively small impact on the cross-lingual signal. Our analysis of attention and tagged decoding provides further supporting evidence for these conclusions.

\section{Future Work}
A natural extension of this work is to investigate a more fine-grained application of tags to both natural and synthetic data, for both back-translation and forward-translation, using quality and domain tags as well as synth-data tags. Similarly, tagging could be investigated as an alternative to data selection, as in \newcite{dynamiccds,Axelrod2011}, or curriculum learning approaches like fine-tuning on in-domain data \cite{W18-6313,model_stacking, DBLP:journals/corr/FreitagA16}. Finally, the token-tagging method should be contrasted with more sophisticated versions of tagging, like concatenating a trainable domain embedding with all token embeddings, as in \newcite{kobus2016controlling}.

\section{Acknowledgements}
Thank you to Markus Freitag, Melvin Johnson and Wei Wang for advising and discussions about these ideas; thank you to Keith Stevens, Mia Chen, and Wei Wang for technical help and bug fixing; thank you to Sergey Edunov for a fast and thorough answer to our question about his paper; and of course to the various people who have given comments and suggestions throughout the process, including Bowen Liang, Naveen Arivazhagan, Macduff Hughes, and George Foster.

\bibliography{tagged_backtranslation}

\begin{thebibliography}{48}
\expandafter\ifx\csname natexlab\endcsname\relax\def\natexlab#1{#1}\fi

\bibitem[{Artetxe et~al.(2018)Artetxe, Labaka, and
  Agirre}]{artetxe2018unsupervised}
Mikel Artetxe, Gorka Labaka, and Eneko Agirre. 2018.
\newblock \href {http://www.aclweb.org/anthology/D18-1399} {{Unsupervised
  Statistical Machine Translation}}.
\newblock In \emph{Proceedings of the 2018 Conference on Empirical Methods in
  Natural Language Processing}, pages 3632--3642.

\bibitem[{Axelrod et~al.(2011)Axelrod, He, and Gao}]{Axelrod2011}
Amittai Axelrod, Xiaodong He, and Jianfeng Gao. 2011.
\newblock {Domain adaptation via pseudo in-domain data selection}.
\newblock In \emph{Proceedings of the 2011 Conference on Empirical Methods in
  Natural Language Processing}, pages 355--362.

\bibitem[{Bahdanau et~al.(2015)Bahdanau, Cho, and Bengio}]{bahdanau2014neural}
Dzmitry Bahdanau, Kyunghyun Cho, and Yoshua Bengio. 2015.
\newblock \href {http://arxiv.org/abs/1409.0473} {{Neural Machine Translation
  by Jointly Learning to Align and Translate}}.
\newblock In \emph{3rd International Conference on Learning Representations,
  {ICLR} 2015}.

\bibitem[{Bertoldi and Federico(2009)}]{bertoldi2009domain}
Nicola Bertoldi and Marcello Federico. 2009.
\newblock {Domain adaptation for statistical machine translation with
  monolingual resources}.
\newblock In \emph{Proceedings of the fourth workshop on statistical machine
  translation}, pages 182--189. Association for Computational Linguistics.

\bibitem[{Bojar and Tamchyna(2011)}]{bojar2011improving}
Ond{\v r}ej Bojar and Ale{\v s} Tamchyna. 2011.
\newblock {Improving translation model by monolingual data}.
\newblock In \emph{Proceedings of the Sixth Workshop on Statistical Machine
  Translation}, pages 330--336. Association for Computational Linguistics.

\bibitem[{Brants et~al.(2007)Brants, Popat, Xu, Och, and
  Dean}]{brants2007large}
Thorsten Brants, Ashok~C Popat, Peng Xu, Franz~J Och, and Jeffrey Dean. 2007.
\newblock \href {https://www.aclweb.org/anthology/D07-1090} {{Large Language
  Models in Machine Translation}}.
\newblock In \emph{Proceedings of the 2007 Joint Conference on Empirical
  Methods in Natural Language Processing and Computational Natural Language
  Learning (EMNLP-CoNLL)}.

\bibitem[{Buduma and Locascio(2017)}]{buduma2017fundamentals}
Nikhil Buduma and Nicholas Locascio. 2017.
\newblock \emph{{Fundamentals of deep learning: Designing next-generation
  machine intelligence algorithms}}.
\newblock ``O'Reilly Media, Inc.".

\bibitem[{Cheng et~al.(2016)Cheng, Xu, He, He, Wu, Sun, and
  Liu}]{cheng2016semisupervised}
Yong Cheng, Wei Xu, Zhongjun He, Wei He, Hua Wu, Maosong Sun, and Yang Liu.
  2016.
\newblock \href {https://www.aclweb.org/anthology/P16-1185} {{Semi-Supervised
  Learning for Neural Machine Translation}}.
\newblock In \emph{Proceedings of the 54th Annual Meeting of the Association
  for Computational Linguistics}, volume~1, pages 1965--1974.

\bibitem[{Cotterell and Kreutzer(2018)}]{cotterell2018explaining}
Ryan Cotterell and Julia Kreutzer. 2018.
\newblock \href {https://arxiv.org/pdf/1806.04402.pdf} {{Explaining and
  Generalizing Back-Translation through Wakesleep}}.
\newblock \emph{arXiv preprint arXiv:1806.04402}.

\bibitem[{Currey et~al.(2017)Currey, Barone, and Heafield}]{currey2017}
Anna Currey, Antonio Valerio~Miceli Barone, and Kenneth Heafield. 2017.
\newblock {Copied monolingual data improves low-resource neural machine
  translation}.
\newblock In \emph{Proceedings of the Second Conference on Machine
  Translation}, pages 148--156.

\bibitem[{Di~He and Ma(2016)}]{he2016dual}
Tao Qin Liwei Wang Nenghai Yu Tieyan~Liu Di~He, Yingce~Xia and Wei-Ying Ma.
  2016.
\newblock \href
  {https://papers.nips.cc/paper/6469-dual-learning-for-machine-translation.pdf}
  {{Dual Learning for Machine Translation}}.
\newblock In \emph{Conference on Advances in Neural Information Processing
  Systems (NeurIPS)}.

\bibitem[{Domhan and Hieber(2017)}]{domhan2017using}
Tobias Domhan and Felix Hieber. 2017.
\newblock \href {https://www.aclweb.org/anthology/D17-1158} {{Using Target-side
  Monolingual Data for Neural Machine Translation through Multi-task
  Learning}}.
\newblock In \emph{Proceedings of the 2017 Conference on Empirical Methods in
  Natural Language Processing}, pages 1500–--1505.

\bibitem[{Edunov et~al.(2018)Edunov, Ott, Auli, and
  Grangier}]{edunov2018understanding}
Sergey Edunov, Myle Ott, Michael Auli, and David Grangier. 2018.
\newblock \href {http://www.aclweb.org/anthology/D18-1045} {{Understanding
  Back-Translation at Scale}}.
\newblock In \emph{Proceedings of the 2018 Conference on Empirical Methods in
  Natural Language Processing (EMNLP)}, pages 489--500.

\bibitem[{Fadaee and Monz(2018)}]{fadaee2018backtranslation}
Marzieh Fadaee and Christof Monz. 2018.
\newblock \href {http://arxiv.org/abs/1808.09006} {{Back-Translation Sampling
  by Targeting Difficult Words in Neural Machine Translation}}.
\newblock \emph{CoRR}, abs/1808.09006.

\bibitem[{Freitag and Al{-}Onaizan(2016)}]{DBLP:journals/corr/FreitagA16}
Markus Freitag and Yaser Al{-}Onaizan. 2016.
\newblock \href {https://arxiv.org/pdf/1612.06897.pdf} {{Fast Domain Adaptation
  for Neural Machine Translation}}.
\newblock \emph{CoRR}, abs/1612.06897.

\bibitem[{Freitag et~al.(2019)Freitag, Caswell, and Roy}]{freitag2019repair}
Markus Freitag, Isaac Caswell, and Scott Roy. 2019.
\newblock \href {http://arxiv.org/abs/1904.04790} {{Text Repair Model for
  Neural Machine Translation}}.
\newblock \emph{CoRR}, abs/1904.04790.

\bibitem[{Garc{\'{\i}}a{-}Mart{\'{\i}}nez
  et~al.(2017)Garc{\'{\i}}a{-}Mart{\'{\i}}nez, {\c C}a{\u g}layan, Aransa,
  Bardet, Bougares, and Barrault}]{garcia2017lium}
Mercedes Garc{\'{\i}}a{-}Mart{\'{\i}}nez, {\" O}zan {\c C}a{\u g}layan, Walid
  Aransa, Adrien Bardet, Fethi Bougares, and Lo{\"{\i}}c Barrault. 2017.
\newblock \href {http://arxiv.org/abs/1707.04499} {{{LIUM} Machine Translation
  Systems for {WMT17} News Translation Task}}.
\newblock \emph{CoRR}, abs/1707.04499.

\bibitem[{Gehring et~al.(2017)Gehring, Auli, Grangier, Yarats, and
  Dauphin}]{gehring2017convolutional}
Jonas Gehring, Michael Auli, David Grangier, Denis Yarats, and Yann~N. Dauphin.
  2017.
\newblock \href {http://dl.acm.org/citation.cfm?id=3305381.3305510}
  {{Convolutional Sequence to Sequence Learning}}.
\newblock In \emph{Proceedings of the 34th International Conference on Machine
  Learning - Volume 70}, pages 1243--1252.

\bibitem[{Gellerstam(1986)}]{gellerstam1986translationese}
Martin Gellerstam. 1986.
\newblock {Translationese in Swedish novels translated from English}.
\newblock \emph{Translation Studies in Scandinavia}, pages 88--95.

\bibitem[{G{\" u}l{\c c}ehre et~al.(2015)G{\" u}l{\c c}ehre, Firat, Xu, Cho,
  Barrault, Lin, Bougares, Schwenk, and Bengio}]{gulcehre2015using}
{\c C}a{\u g}lar G{\" u}l{\c c}ehre, Orhan Firat, Kelvin Xu, Kyunghyun Cho,
  Loic Barrault, Huei-Chi Lin, Fethi Bougares, Holger Schwenk, and Yoshua
  Bengio. 2015.
\newblock \href {https://arxiv.org/abs/1503.03535} {{On using Monolingual
  Corpora in Neural Machine Translation}}.
\newblock \emph{arXiv preprint arXiv:1503.03535}.

\bibitem[{G{\" u}l{\c c}ehre et~al.(2017)G{\" u}l{\c c}ehre, Firat, Xu, Cho,
  and Bengio}]{gulcehre2017integrating}
{\c C}a{\u g}lar G{\" u}l{\c c}ehre, Orhan Firat, Kelvin Xu, Kyunghyun Cho, and
  Yoshua Bengio. 2017.
\newblock \href {https://doi.org/10.1016/j.csl.2017.01.014} {{On Integrating a
  Language Model into Neural Machine Translation}}.
\newblock \emph{Comput. Speech Lang.}, pages 137--148.

\bibitem[{Ha et~al.(2017)Ha, Niehues, and Waibel}]{ha2017effective}
Thanh{-}Le Ha, Jan Niehues, and Alexander~H. Waibel. 2017.
\newblock \href {http://arxiv.org/abs/1711.07893} {{Effective Strategies in
  Zero-Shot Neural Machine Translation}}.
\newblock \emph{CoRR}, abs/1711.07893.

\bibitem[{{Hassan Sajjad and Nadir Durrani and Fahim Dalvi and Yonatan Belinkov
  and Stephan Vogel}(2017)}]{model_stacking}
{Hassan Sajjad and Nadir Durrani and Fahim Dalvi and Yonatan Belinkov and
  Stephan Vogel}. 2017.
\newblock Neural machine translation training in a multi-domain scenario.
\newblock \emph{arXiv preprint arXiv:1708.08712v2}.

\bibitem[{He et~al.(2016)He, He, Wu, and Wang}]{he2016improved}
Wei He, Zhongjun He, Hua Wu, and Haifeng Wang. 2016.
\newblock {Improved neural machine translation with SMT features}.
\newblock In \emph{Thirtieth AAAI conference on artificial intelligence}.

\bibitem[{Imamura et~al.(2018)Imamura, Fujita, and
  Sumita}]{imamura2018enhancement}
Kenji Imamura, Atsushi Fujita, and Eiichiro Sumita. 2018.
\newblock \href {http://aclweb.org/anthology/W18-2707} {{Enhancement of Encoder
  and Attention Using Target Monolingual Corpora in Neural Machine
  Translation}}.
\newblock In \emph{Proceedings of the 2nd Workshop on Neural Machine
  Translation and Generation}, volume~1, pages 55–--63.

\bibitem[{Johnson et~al.(2016)Johnson, Schuster, Le, Krikun, Wu, Chen, Thorat,
  Vi{'{e}}gas, Wattenberg, Corrado, Hughes, and Dean}]{johnson2016google}
Melvin Johnson, Mike Schuster, Quoc~V. Le, Maxim Krikun, Yonghui Wu, Zhifeng
  Chen, Nikhil Thorat, Fernanda~B. Vi{'{e}}gas, Martin Wattenberg, Greg
  Corrado, Macduff Hughes, and Jeffrey Dean. 2016.
\newblock \href {http://arxiv.org/abs/1611.04558} {{Google's Multilingual
  Neural Machine Translation System: Enabling Zero-Shot Translation}}.
\newblock \emph{CoRR}, abs/1611.04558.

\bibitem[{Kim and Rush(2016)}]{kim-rush-2016-sequence}
Yoon Kim and Alexander~M. Rush. 2016.
\newblock \href {https://doi.org/10.18653/v1/D16-1139} {{Sequence-Level
  Knowledge Distillation}}.
\newblock In \emph{Proceedings of the 2016 Conference on Empirical Methods in
  Natural Language Processing}, pages 1317--1327, Austin, Texas. Association
  for Computational Linguistics.

\bibitem[{Kobus et~al.(2016)Kobus, Crego, and Senellart}]{kobus2016controlling}
Catherine Kobus, Josep~Maria Crego, and Jean Senellart. 2016.
\newblock \href {http://arxiv.org/abs/1612.06140} {{Domain Control for Neural
  Machine Translation}}.
\newblock \emph{CoRR}, abs/1612.06140.

\bibitem[{Koehn(2009)}]{koehn2009statistical}
Philipp Koehn. 2009.
\newblock \emph{{Statistical machine translation}}.
\newblock Cambridge University Press.

\bibitem[{Kuczmarski and Johnson(2018)}]{kuczmarski2018gender}
James Kuczmarski and Melvin Johnson. 2018.
\newblock \href {https://www.tdcommons.org/dpubs_series/1577} {Gender-aware
  natural language translation}.
\newblock \emph{Technical Disclosure Commons}.

\bibitem[{Lambert et~al.(2011)Lambert, Schwenk, Servan, and
  Abdul-Rauf}]{lambert2011investigations}
Patrik Lambert, Holger Schwenk, Christophe Servan, and Sadaf Abdul-Rauf. 2011.
\newblock {Investigations on translation model adaptation using monolingual
  data}.
\newblock In \emph{Proceedings of the Sixth Workshop on Statistical Machine
  Translation}, pages 284--293. Association for Computational Linguistics.

\bibitem[{Lample and Conneau(2019)}]{lample2019cross}
Guillaume Lample and Alexis Conneau. 2019.
\newblock {Cross-lingual Language Model Pretraining}.
\newblock \emph{arXiv preprint arXiv:1901.07291}.

\bibitem[{Lample et~al.(2018{\natexlab{a}})Lample, Conneau, Denoyer, and
  Ranzato}]{lample2017unsupervised}
Guillaume Lample, Alexis Conneau, Ludovic Denoyer, and Marc'Aurelio Ranzato.
  2018{\natexlab{a}}.
\newblock \href {https://openreview.net/forum?id=rkYTTf-AZ} {{Unsupervised
  Machine Translation Using Monolingual Corpora Only}}.
\newblock In \emph{International Conference on Learning Representations}.

\bibitem[{Lample et~al.(2018{\natexlab{b}})Lample, Ott, Conneau, Denoyer, and
  Ranzato}]{lample2018phrase}
Guillaume Lample, Myle Ott, Alexis Conneau, Ludovic Denoyer, and Marc'Aurelio
  Ranzato. 2018{\natexlab{b}}.
\newblock \href {http://aclweb.org/anthology/D18-1549} {{Phrase-Based \& Neural
  Unsupervised Machine Translation}}.
\newblock In \emph{Proceedings of the 2018 Conference on Empirical Methods in
  Natural Language Processing (EMNLP)}.

\bibitem[{Niu et~al.(2018)Niu, Denkowski, and Carpuat}]{niu2018bidirectional}
Xing Niu, Michael Denkowski, and Marine Carpuat. 2018.
\newblock \href {https://www.aclweb.org/anthology/W18-27#page=96}
  {{Bi-Directional Neural Machine Translation with Synthetic Parallel Data}}.
\newblock \emph{ACL 2018}, page~84.

\bibitem[{Papineni et~al.(2002)Papineni, Roukos, Ward, and
  Zhu}]{papineni2002bleu}
Kishore Papineni, Salim Roukos, Todd Ward, and Wei-Jing Zhu. 2002.
\newblock \href {https://www.aclweb.org/anthology/P02-1040} {{BLEU: a Method
  for Automatic Evaluation of Machine Translation}}.
\newblock In \emph{Proceedings of the 40th annual meeting on association for
  computational linguistics}, pages 311--318. Association for Computational
  Linguistics.

\bibitem[{Poncelas et~al.(2018)Poncelas, Shterionov, Way, de~Buy~Wenniger, and
  Passban}]{poncelas2018investigating}
Alberto Poncelas, Dimitar Shterionov, Andy Way, Gideon~Maillette
  de~Buy~Wenniger, and Peyman Passban. 2018.
\newblock \href {https://dialnet.unirioja.es/servlet/articulo?codigo=6474368}
  {{Investigating Backtranslation in Neural Machine Translation}}.
\newblock In \emph{Proceedings of the 21st Annual Conference of the European
  Association for Machine Translation}, pages 249--258.

\bibitem[{Post(2018)}]{post2018call}
Matt Post. 2018.
\newblock \href {https://arxiv.org/abs/1804.08771} {{A Call for Clarity in
  Reporting Bleu Scores}}.
\newblock \emph{arXiv preprint arXiv:1804.08771}.

\bibitem[{Sennrich et~al.(2017)Sennrich, Birch, Currey, Germann, Haddow,
  Heafield, Barone, and Williams}]{sennrich2017edinburgh}
Rico Sennrich, Alexandra Birch, Anna Currey, Ulrich Germann, Barry Haddow,
  Kenneth Heafield, Antonio Valerio~Miceli Barone, and Philip Williams. 2017.
\newblock \href {http://arxiv.org/abs/1708.00726} {{The University of
  Edinburgh's Neural {MT} Systems for {WMT17}}}.
\newblock \emph{CoRR}, abs/1708.00726.

\bibitem[{Sennrich et~al.(2016)Sennrich, Haddow, and
  Birch}]{sennrich2016improving}
Rico Sennrich, Barry Haddow, and Alexandra Birch. 2016.
\newblock \href {http://www.aclweb.org/anthology/P16-1009} {{Improving Neural
  Machine Translation Models with Monolingual Data}}.
\newblock In \emph{Proceedings of the 54th Annual Meeting of the Association
  for Computational Linguistics (Volume 1: Long Papers)}, pages 86--96.

\bibitem[{Shen et~al.(2019)Shen, Nguyen, Wu, Chen, Chen, Jia, Kannan, Sainath,
  and et~al.}]{shen2019lingvo}
Jonathan Shen, Patrick Nguyen, Yonghui Wu, Zhifeng Chen, Mia~X. Chen, Ye~Jia,
  Anjuli Kannan, Tara~N. Sainath, and Yuan~Cao et~al. 2019.
\newblock \href {http://arxiv.org/abs/1902.08295} {{Lingvo: a Modular and
  Scalable Framework for Sequence-to-Sequence Modeling}}.
\newblock \emph{CoRR}, abs/1902.08295.

\bibitem[{Thompson et~al.(2018)Thompson, Khayrallah, Anastasopoulos, McCarthy,
  Duh, Marvin, McNamee, Gwinnup, Anderson, and Koehn}]{W18-6313}
Brian Thompson, Huda Khayrallah, Antonios Anastasopoulos, Arya~D. McCarthy,
  Kevin Duh, Rebecca Marvin, Paul McNamee, Jeremy Gwinnup, Tim Anderson, and
  Philipp Koehn. 2018.
\newblock \href {http://aclweb.org/anthology/W18-6313} {{Freezing Subnetworks
  to Analyze Domain Adaptation in Neural Machine Translation}}.
\newblock In \emph{Proceedings of the Third Conference on Machine Translation:
  Research Papers}, pages 124--132. Association for Computational Linguistics.

\bibitem[{Ueffing et~al.(2007)Ueffing, Haffari, and Sarkar}]{ueffing2007semi}
Nicola Ueffing, Gholamreza Haffari, and Anoop Sarkar. 2007.
\newblock {Semi-supervised model adaptation for statistical machine
  translation}.
\newblock \emph{Machine Translation}, 21(2):77--94.

\bibitem[{Vaswani et~al.(2017)Vaswani, Shazeer, Parmar, Uszkoreit, Jones,
  Gomez, Kaiser, and Polosukhin}]{vaswani2017attention}
Ashish Vaswani, Noam Shazeer, Niki Parmar, Jakob Uszkoreit, Llion Jones,
  Aidan~N Gomez, {\L}ukasz Kaiser, and Illia Polosukhin. 2017.
\newblock \href
  {https://papers.nips.cc/paper/7181-attention-is-all-you-need.pdf} {{Attention
  Is All You Need}}.
\newblock In \emph{{Advances in Neural Information Processing Systems}}, pages
  5998--6008.

\bibitem[{Vu~Cong Duy~Hoang and Cohn(2018)}]{hoang2018iterative}
Gholamreza~Haffari Vu~Cong Duy~Hoang, Philipp~Koehn and Trevor Cohn. 2018.
\newblock \href {http://www.aclweb.org/anthology/W18-2703} {{Iterative
  Backtranslation for Neural Machine Translation}}.
\newblock In \emph{Proceedings of the 2nd Workshop on Neural Machine
  Translation and Generation}, volume~1, pages 18--24.

\bibitem[{van~der Wees et~al.(2017)van~der Wees, Bisazza, and
  Monz}]{dynamiccds}
Marlies van~der Wees, Arianna Bisazza, and Christof Monz. 2017.
\newblock {Dynamic Data Selection for Neural Machine Transaltion}.
\newblock In \emph{Proceedings of the 2017 Conference on Empirical Methods in
  Natural Language Processing}, pages 1400--1410.

\bibitem[{Xia et~al.(2017)Xia, Qin, Chen, Bian, Yu, and Liu}]{xia2017dual}
Yingce Xia, Tao Qin, Wei Chen, Jiang Bian, Nenghai Yu, and Tie-Yan Liu. 2017.
\newblock \href {https://arxiv.org/pdf/1707.00415.pdf} {{Dual Supervised
  Learning}}.
\newblock In \emph{International Conference on Machine Learning (ICML)}.

\bibitem[{Yamagishi et~al.(2016)Yamagishi, Kanouchi, Sato, and
  Komachi}]{yamagishi2016controlling}
Hayahide Yamagishi, Shin Kanouchi, Takayuki Sato, and Mamoru Komachi. 2016.
\newblock {Controlling the voice of a sentence in Japanese-to-English neural
  machine translation}.
\newblock In \emph{Proceedings of the 3rd Workshop on Asian Translation
  (WAT2016)}, pages 203--210.

\end{thebibliography}
\bibliographystyle{acl_natbib}

\end{document}